\documentclass[10pt,twocolumn,letterpaper]{article}

\usepackage{cvpr}              

\usepackage{mathtools}
\usepackage{url}
\usepackage{booktabs} 
\usepackage{array} %
\usepackage{multirow}
\usepackage{graphicx}
\usepackage{amsmath, amssymb, amsthm}
\usepackage{booktabs} 
\usepackage{graphicx} 
\usepackage{mathtools}   
\usepackage{array}

\usepackage{booktabs}
\usepackage{multirow}
\usepackage{makecell}
\usepackage{xcolor,colortbl}
\usepackage{xurl}
\usepackage{float}
\usepackage{algorithm, algorithmic}
\usepackage{xcolor,colortbl}

\usepackage{booktabs}
\usepackage{xcolor}
\usepackage{multirow}
\usepackage{graphicx}

\usepackage{booktabs}
\usepackage{xcolor}
\usepackage{multirow}
\usepackage{graphicx}

\newcommand{\name}{\emph{\textbf{PoInit-of-View}}}

\newcommand{\poisoned}[1]{\textcolor{red!65!black}{(#1)}} 

\usepackage{tikz}
\usepackage{pgfplots}
\pgfplotsset{compat=newest}

\definecolor{cvprblue}{rgb}{0.21,0.49,0.74}
\usepackage[pagebackref,breaklinks,colorlinks,allcolors=cvprblue]{hyperref}

\def\paperID{****} 
\def\confName{CVPR}
\def\confYear{2026}

\title{PoInit-of-View: Poisoning Initialization of Views\\Transfers Across Multiple 3D Reconstruction Systems}
\author{
    Weijie Wang$^{1,2}$ \quad 
    Songlong Xing$^{1}$ \quad 
    Zhengyu Zhao$^{3}$\thanks{Corresponding author.}\quad
    Nicu Sebe$^{1}$ \quad 
    Bruno Lepri$^{2}$ \\
    $^1$University of Trento, Italy \quad 
    $^2$Fondazione Bruno Kessler, Italy \quad 
    $^3$Xi'an Jiaotong University, China \\
    {\tt\small \{weijie.wang,songlong.xing,niculae.sebe\}@unitn.it, zhengyu.zhao@xjtu.edu.cn}
}

\usepackage{amsmath,amsfonts,bm}

\def\eqref#1{equation~\ref{#1}}

\def\1{\bm{1}}

\DeclareMathAlphabet{\mathsfit}{\encodingdefault}{\sfdefault}{m}{sl}
\SetMathAlphabet{\mathsfit}{bold}{\encodingdefault}{\sfdefault}{bx}{n}

\begin{document}
\maketitle
\begin{abstract}
Poisoning input views of 3D reconstruction systems has been recently studied.
However, we identify that existing studies simply backpropagate adversarial gradients through the 3D reconstruction pipeline as a whole, without uncovering the new vulnerability rooted in specific modules of the 3D reconstruction pipeline.
In this paper, we argue that the structure-from-motion (SfM) initialization, as the geometric core of many widely used reconstruction systems, can be targeted to achieve transferable poisoning effects across diverse 3D reconstruction systems. 
To this end, we propose \name, which optimizes adversarial perturbations to intentionally introduce cross-view gradient inconsistencies at projections of corresponding 3D points. 
These inconsistencies disrupt keypoint detection and feature matching, thereby corrupting pose estimation and triangulation within SfM, eventually resulting in low-quality rendered views.  
We also provide a theoretical analysis that connects cross-view inconsistency to correspondence collapse.
Experimental results demonstrate the effectiveness of our \name~on diverse 3D reconstruction systems and datasets, surpassing the single-view baseline by 25.1\% in PSNR and 16.5\% in SSIM in black-box transfer settings, such as 3DGS to NeRF.

\end{abstract}


\section{Introduction}
3D reconstruction has experienced a surge of innovation, becoming one of the most active research areas in computer vision~\citep{barron2022mipnerf360, nie2024t2td, chen2024sgnerf,wang2024uvmap,li2025freeinsert, li2025cross,li2026token}. 
Driven by the rise of multi-view imaging and neural implicit representations, 3D reconstruction has been deployed in a wide range of safety-critical applications, including autonomous driving~\cite{Jiang_2023_ICCV, Jia_2023_ICCV}, AR/VR~\cite{Yang_2025_CVPR, Luo_2024_CVPR}, robotics navigation~\cite{Yugay_2025_CVPR, wang2025fully}, and high-precision guidance in robot-assisted surgery~\cite{li2025robotic_visual_instruction}. 
Given their high-stakes nature, any disruption or malicious manipulation can introduce errors that propagate downstream and lead to catastrophic consequences. 
Therefore, understanding and mitigating their adversarial risks is crucial.

\begin{figure}[!t]
    \centering
    \includegraphics[width=0.9\linewidth]{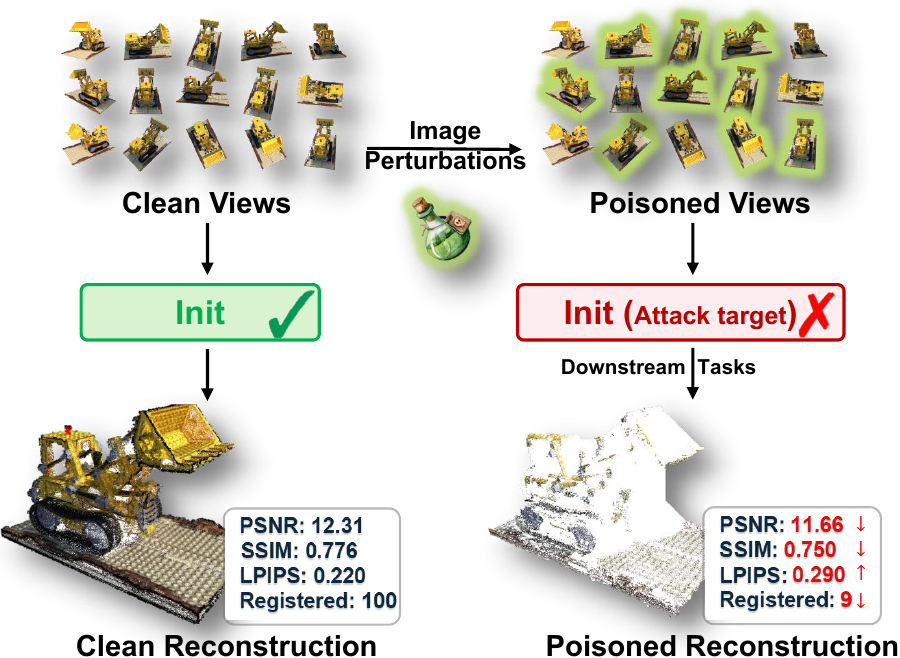}
    \caption{
    Our \name~method injects imperceptible perturbations into input views (top) to induce cross-view inconsistency (see details in \Cref{fig:epipolar_constraint}) in the foundational Structure-from-Motion (SfM) initialization rather than the single-view-based loss in existing attacks.
    The corrupted feature correspondences propagate to inaccurate point triangulation, leading to erroneous camera poses and severely degraded 3D reconstruction results (bottom).
    Compared to clean views, our poisoned inputs yield far fewer registered images, and as a result, lower quality of rendered views.
    }
    \label{fig:teaser}
\end{figure}

There are already several attempts~\cite{lin2023nerfool,horvath2023targeted,wang2024benchmarking,wang2023turn,song2024geometry,ke2025stealthattack,meng2025il2nerf} to expose the adversarial vulnerabilities of 3D vision, particularly 3D reconstruction with Neural Radiance Fields (NeRF) and 3D Gaussian Splatting (3DGS). 
However, we identify that these attempts simply backpropagate adversarial gradients through the 3D reconstruction pipeline as a whole, without uncovering the new vulnerability rooted in specific modules of the 3D reconstruction pipeline.

In this paper, we argue that the initialization module, as the geometric core of the reconstruction pipeline, is especially vulnerable to adversarial perturbations.
This initialization, typically implemented via structure-from-motion (SfM)~\citep{Ullman1979}, performs keypoint detection, feature matching, and camera pose estimation to establish a sparse 3D structure, which in turn provides the geometric backbone for subsequent dense depth estimation and surface modeling~\cite{schoenberger2016sfm, schonberger2016pixelwise, mueller2022instant, kerbl3Dgaussians, chen2024sgnerf}. 
Because SfM has been widely adopted in most modern 3D reconstruction and neural rendering systems, such as MVS~\cite{schoenberger2016sfm, schonberger2016pixelwise}, NeRF~\cite{mildenhall2021nerf}, and 3DGS~\cite{kerbl3Dgaussians}, poisoning on SfM would disrupt the foundation of 3D reconstruction and naturally propagate the adversarial effects to subsequent modules.
This also makes it possible to achieve highly transferable perturbations~\cite{liu2017delving,dong2018boosting,zhao2021success,zhao2025revisit}, because the perturbations do not overfit to specific reconstruction architectures.

To study the above important vulnerability of 3D reconstruction, we propose \textbf{PoInit-of-View}, a \textbf{Po}isoning method targeting the \textbf{Init}ialization of input \textbf{View}s.
As illustrated in ~\Cref{fig:teaser}, it injects imperceptible perturbations into a subset of multi-view input images to corrupt local feature correspondences and to induce cross-view gradient and photometric inconsistencies.
These inconsistencies propagate through triangulation and pose estimation, causing SfM to register substantially fewer cameras and subsequently produce fewer triangulated points. 
Such an insufficient number of triangulated points would render it impossible to reconstruct meaningful 3D space and thus fail to produce high-quality views (See Theorem 1 for why cross-view inconsistency leads to correspondence collapse in SfM).
In contrast, existing work is limited to disrupting features within a single view but not involving cross-view interactions.
Our contributions are summarized as follows:
\begin{itemize}
    \item We identify a fundamental adversarial vulnerability in 3D reconstruction, showing that the geometric initialization in the widely-adopted Structure-from-Motion (SfM) module is intrinsically vulnerable to subtle cross-view inconsistencies, which can lead to system-level failures of the entire 3D reconstruction pipeline.
    \item We propose \name, which poisons the initialization of a subset of input image views to induce cross-view inconsistencies in SfM. Such inconsistencies would subsequently affect the following rendering process and eventually result in low-quality rendered views.
    \item We demonstrate that \name, can achieve strong adversarial effects across diverse reconstruction systems (e.g., MVS, NeRF, and 3DGS) on multiple datasets, surpassing the single-view baseline by 25.1\% in PSNR and 16.5\% in SSIM in black-box transfer (e.g., 3DGS to NeRF) settings.
\end{itemize}

\section{Related Work}

\subsection{3D Reconstruction and Novel-View Synthesis}
Most existing 3D reconstruction and novel-view synthesis techniques rely on accurate camera intrinsics and extrinsics, which are typically pre-estimated and calibrated using SfM (Structure-from-Motion) or similar multi-view geometry tools (e.g., COLMAP)~\citep{schoenberger2016sfm}.
\textbf{Multi-view reconstruction networks} (MVSNet~\citep{yao2018mvsnet}, UniMVSNet~\citep{unimvsnet}, NeuralRecon~\citep{sun2021neuralrecon}, MVSTER~\citep{wang2022mvster}) depend on per-image calibrations to build cost volumes or update TSDFs.
\textbf{Layer- and voxel-based renderers} MPI~\citep{zhou2018stereo}, Structural-MPI~\citep{zhang2023structuralmpi}, DeepVoxels~\citep{sitzmann2019deepvoxels}, Vgos~\citep{sun2023vgos} require known camera matrices to reproject 3D features onto the image plane.
\textbf{Implicit neural renderers}, including NeRF~\citep{mildenhall2021nerf}, BARF~\citep{lin2021barf}, and UnMix-NeRF~\citep{perez2025unmix} assume SfM-provided coarse pose alignments as initialization for optimization.
More recently, \textbf{3D Gaussian Splatting}~\citep{kerbl3Dgaussians}, Wild‑GS~\citep{xu2024wild}, and FewViewGS~\citep{yin2024fewviewgs} have achieved efficient real-time synthesis by leveraging explicit Gaussian primitives and splatting-based rendering. 
These systems still rely heavily on camera poses derived from SfM to accurately reconstruct 3D scenarios.
 
\subsection{Adversarial Attacks on 3D Reconstruction}
Existing attacks on 3D Reconstruction mainly focus on NeRF and 3D Gaussian Splatting (3DGS).
For NeRF, \textit{NeRFool}~\citep{lin2023nerfool} and \textit{IL2-NeRF}~\citep{meng2025il2nerf} degrade novel-view synthesis quality under white-box settings, requiring access to model losses and camera parameters.
For 3DGS, \textit{Poison-Splat}~\citep{lu2025poisonsplat} poisons the densification process to cause primitive explosion and excessive memory usage (a denial-of-service effect), while \textit{GaussTrap}~\citep{hong2025gausstrapstealthypoisoningattacks} embeds stealthy backdoors to induce targeted rendering failures.
Other studies \cite{horvath2023targeted,song2024geometry,ke2025stealthattack} focus on generating rendered views with specific natural semantics.
There are also studies using 3D techniques to craft 2D or 3D adversarial examples for fooling downstream recognition models~\citep{dong2022viewfool,li2023adv3d,zeybey2024gaussian,zheng2024physical,jiang2024nerfail,jiang2025mpam}, which are out of our scope.
The above attacks have not uncovered new vulnerability rooted in specific modules of the 3D reconstruction pipeline, but our \name~ targets the fundamental vulnerability in the geometric core: the initialization module of many widely used 3D reconstruction systems.

\section{Method}

\begin{figure*}[t]
    \centering
    \includegraphics[width=\linewidth]{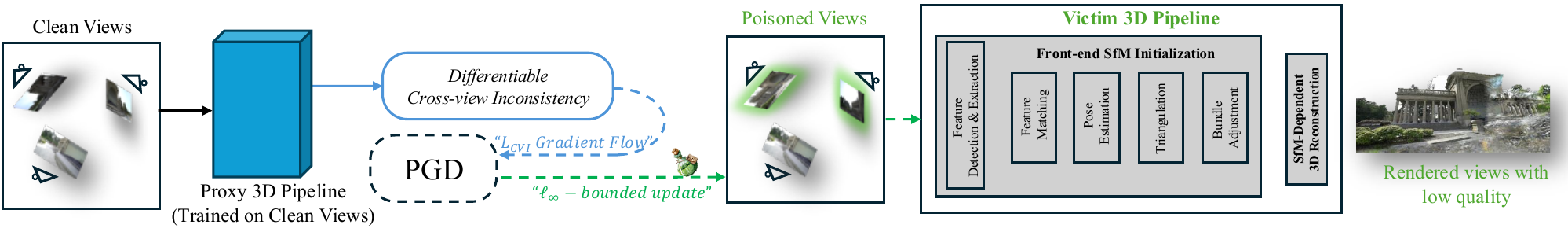}
    \caption{Framework of PoInit-of-View.}
    \label{fig:placeholder}
\end{figure*}

In this section, we introduce \name, a new poisoning method that aims to induce cross-view inconsistencies in the SfM initialization module to disrupt the rendered views of 3D reconstruction.
~\Cref{fig:placeholder} gives an overview of our \name.

\subsection{Preliminary}
\paragraph{SfM as the Basis for 3D Reconstruction.}
NeRF~\citep{mildenhall2021nerf} and 3D Gaussian Splatting (3DGS)~\citep{kerbl3Dgaussians} achieve high-fidelity view synthesis and 3D reconstruction. 
Although their internal scene representations differ (NeRF models volumetric radiance and density while 3DGS uses anisotropic Gaussian primitives), both pipelines typically rely on a preceding Structure-from-Motion (SfM) stage~\citep{schoenberger2016sfm} for initialization.
SfM provides camera intrinsics and extrinsics together with a sparse 3D point cloud that form the geometric scaffold for subsequent dense reconstruction or neural optimization; without a reliable SfM initialization, downstream optimizers often fail to converge to a coherent scene.

Formally, let $\{I_i\}_{i=1}^N$ denote the input views. For image $i$, we denote keypoints by $p_{i,k} \in \mathbb{R}^2$ and their associated descriptors by $\phi(p_{i,k}) \in \mathbb{R}^d$.
Candidate correspondences between a keypoint $p_{i,k}$ in image $i$ and $p_{j,\ell}$ in image $j$ are established in descriptor space (e.g., nearest neighbor with Lowe's ratio test and mutual check) and geometrically verified via RANSAC using the epipolar constraint:
\begin{equation}
    p_{j,\ell}^{\top} \mathbf{F}_{ij}\, p_{i,k} = 0,
\end{equation}
where $\mathbf{F}_{ij}$ is the fundamental matrix between images $i$ and $j$.
If camera intrinsics $\mathbf{K}_i, \mathbf{K}_j$ are known, one may work with the essential matrix $\mathbf{E}_{ij} = \mathbf{K}_j^{\top} \mathbf{F}_{ij}\mathbf{K}_i$ and recover the relative pose $(\mathbf{R}_{ij}, \mathbf{t}_{ij})$ via the five-/eight-point algorithm.

Given validated correspondences and initial relative poses, SfM jointly refines global camera poses $\{(\mathbf{R}_i,\mathbf{t}_i)\}$ and 3D points $\{\mathbf{X}_j\}$ by minimizing the robust reprojection error:
\begin{equation}
\min_{\{\mathbf{R}_i,\mathbf{t}_i\},\,\{\mathbf{X}_j\}}
  \sum_{i,j} \rho\!\left(\|p_{ij}-\pi(\mathbf{K}_i,\mathbf{R}_i,\mathbf{t}_i,\mathbf{X}_j)\|_2^2\right),
\label{eq:sfm}
\end{equation}
where $p_{ij}$ denotes the observed image location of 3D point $\mathbf{X}_j$ in image $i$, $\pi(\cdot)$ is the projection model, and $\rho(\cdot)$ is a robust penalty (e.g., Huber).
This initialization is indispensable for many downstream methods: NeRF relies on accurate poses for photometric optimization, while 3DGS uses sparse SfM points as anchors for Gaussian placement and densification.
Our attack focuses solely on the SfM front-end (keypoint detection, descriptor matching, pose estimation, and triangulation); it does not require access to NeRF or 3DGS gradients and is therefore downstream-agnostic.
SfM assumes approximately stable local appearance and sufficient overlap across views; our attack deliberately violates this assumption by inducing cross-view gradient inconsistencies that disrupt descriptor matching and destabilize the entire reconstruction pipeline.

\subsection{Problem Statement}

\paragraph{Attacker.}
Given a clean multi-view image set $\mathcal{I}$ and a perceptual budget $\varepsilon$, the attacker aims to synthesize a poisoned set $\tilde{\mathcal{I}}$ satisfying $\|\tilde{\mathcal{I}}-\mathcal{I}\|_{\infty}\le\varepsilon$, such that an off-the-shelf SfM pipeline (e.g., COLMAP) fed with $\tilde{\mathcal{I}}$ produces a \emph{significant degradation} in geometric initialization, i.e., a sharp drop in the number of registered images, triangulated keypoints, and total sparse 3D points, ultimately causing downstream reconstruction (MVS/NeRF/3DGS) to fail.

\paragraph{Threat Model (Black-Box) and Proxy.}
The attacker has \emph{black-box} access to the victim SfM service: they can upload images and observe reconstruction outputs (e.g., registered images, triangulated keypoints, total 3D points) but have no access to internal descriptors, thresholds, or source code.
To compute gradients for optimization, the attacker uses a locally available differentiable \emph{proxy} model that provides a tractable surrogate loss (e.g., the cross-view gradient inconsistency loss $\mathcal{L}_{\mathrm{CVI}}$ as Eq. \ref{eq:cvi}).
The proxy is used only for optimization; all evaluations and reported results use the unmodified, off-the-shelf SfM pipeline.

\paragraph{Optimization Objective.}
Directly minimizing SfM success probability is intractable, so the attacker solves a differentiable surrogate problem:
\begin{equation}
\scriptsize
    \max_{\tilde{\mathcal{I}}} 
    \mathcal{L}_{\mathrm{CVI}}(\tilde{\mathcal{I}}) 
    - \lambda_{\mathrm{SSIM}}\big(1-\mathrm{SSIM}(\tilde{\mathcal{I}},\mathcal{I})\big)
    - \lambda_{\mathrm{TV}}\mathrm{TV}(\tilde{\mathcal{I}}),
    \quad 
    \text{s.t. } \|\tilde{\mathcal{I}}-\mathcal{I}\|_\infty \le \varepsilon,
\label{eq:attack_obj}
\end{equation}
where $\mathcal{L}_{\mathrm{CVI}}$ encourages cross-view inconsistency, and the regularization terms ensure visual imperceptibility and smoothness. The optimization is performed via projected gradient ascent (PGD), guided by gradients through the differentiable proxy.

\paragraph{Victim.}
The victim receives a dataset $\mathcal{D}_p = \{\tilde{I}_k\}_{k=1}^N$ and, unaware of any poisoning, runs a standard SfM pipeline to recover camera poses $\{(\mathbf{R}_i,\mathbf{t}_i)\}$ and sparse points $\{\mathbf{X}_j\}$ by minimizing the reprojection error in Eq.~\ref{eq:sfm}.
The goal of the victim is to obtain a consistent geometric initialization (high inlier ratios, well-conditioned pose graph, accurate sparse points), which downstream modules rely upon.
Because our attack corrupts this initialization, degradation in keypoint matching or inlier statistics directly undermines downstream reconstruction.
We quantify attack impact via internal SfM statistics: registered images, triangulated keypoints, and total sparse 3D points.

\begin{figure}[!t]
    \centering
    \includegraphics[width=0.9\linewidth]{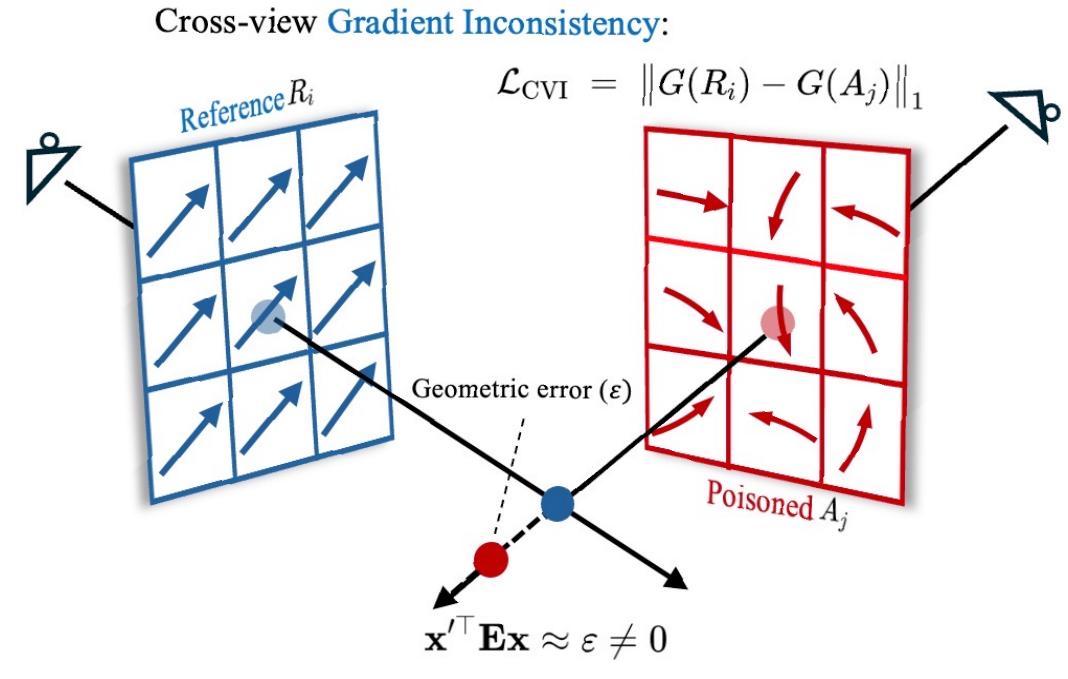}
    \caption{Cross-view gradient inconsistency causes geometric projection error.
    For a 3D point $X$, inconsistent local gradients between reference $R_i$ and poisoned $A_j$ increase
    $L_{\text{CVI}} = \|G(R_i)-G(A_j)\|_1$, leading to mismatched correspondences that violate the epipolar constraint
    $\mathbf{x}'^{\top}\mathbf{E}\mathbf{x} \approx \varepsilon \neq 0$.}
    \label{fig:epipolar_constraint}
\end{figure}

\paragraph{Multi-view geometric consistency.}
Structure-from-Motion (SfM) recovers camera poses by enforcing
cross-view feature correspondences to satisfy a geometric constraint.
Given a pair of calibrated views with corresponding pixels
$\mathbf{x}$ and $\mathbf{x}'$, their relation follows the epipolar constraint: $\mathbf{x}'^{\top} \mathbf{E}\, \mathbf{x} = 0$, 
where $\mathbf{E} = [\mathbf{t}]_{\times}\mathbf{R}$ is the essential matrix derived
from relative rotation $\mathbf{R}$ and translation $\mathbf{t}$.
Small appearance changes that disturb this relation
lead to mismatched correspondences and unstable pose estimation.
As shown in \Cref{fig:epipolar_constraint}, our attack targets this front-end dependency by introducing
cross-view inconsistency, causing $\mathbf{x}'^{\top}\mathbf{E}\mathbf{x}\approx \varepsilon \ne 0$, which leads to erroneous triangulation.

\subsection{Cross-View Gradient Inconsistency}
\label{sec:method:theory}

\paragraph{Notation.}
For an image $I_i : \Omega \subset \mathbb{R}^2 \!\to\! [0,1]$, we denote its Sobel gradient as 
$G(I_i) = (\partial_x I_i, \partial_y I_i)$.
Let $\phi : \mathbb{R}^{2\times|\Omega|} \!\to\! \mathbb{R}^{128}$ be a local descriptor (e.g., SIFT~\cite{lowe2004distinctive}) applied to a patch $P_r$ of radius $r$ around a keypoint.

\noindent\textbf{Assumption 1 (Cross-View Gradient Consistency).}
For projections of the same 3D point onto two clean views $(I_i, I_j)$, the local gradients satisfy
\begin{equation}
    \|G(I_i(p_i)) - G(I_j(p_j))\|_1 \le \tau_g.
    \label{eq:cgc}
\end{equation}
This reflects the empirical fact that corresponding regions across views exhibit similar edge and texture structure, a property fundamental to descriptor repeatability.

\noindent\textbf{Definition 1 (Cross-View Inconsistency Loss).}
Given a reference view $R_i$ and a poison-target view $A_j$, we optimize
\begin{equation}
    \mathcal{L}_{\mathrm{CVI}} = 
    \|G(R_i) - G(A_j)\|_1,
\label{eq:cvi}
\end{equation}
optionally averaged over multiple view pairs in practice.

\noindent\textbf{Assumption 2 (Local Lipschitz Continuity).}
There exist constants $L_r^{\min}, L_r^{\max}>0$ such that for any two patches of radius $r$:
\begin{equation}
\footnotesize
L_r^{\min} \|G_1-G_2\|_1
\le \|\phi(G_1)-\phi(G_2)\|_2
\le L_r^{\max}\|G_1-G_2\|_1.
\label{eq:lipschitz}
\end{equation}
This assumption is supported empirically: both SIFT histograms and CNN-based descriptors with spectral normalization exhibit bounded sensitivity to gradient perturbations.

\paragraph{From Gradient Inconsistency to Feature Mismatch.}
\textbf{Lemma 1 (Descriptor Distance Bound).}
If $\mathcal{L}_{\mathrm{CVI}} \ge \tau_g+\Delta$ for some $\Delta>0$, then local descriptors deviate by
\begin{equation}\footnotesize
\|\phi(G(R_i))-\phi(G(A_j))\|_2 \ge \beta_r \Delta,
\quad \text{where } \beta_r = L_r^{\min}.
\end{equation}
Thus, exceeding the gradient-consistency bound forces the corresponding descriptors to diverge proportionally.

\noindent\textbf{Assumption 3 (Light-Tailed Descriptor Distribution).}
Descriptor deviations follow a sub-exponential tail:
\begin{equation}
\Pr[\|\phi(G(R_i))-\phi(G(A_j))\|_2 < \tau_d]
\le \exp(-\alpha \,\beta_r \Delta)
\end{equation}
for some $\alpha>0$, a standard modeling choice also used in robustness analyses of SIFT-like features.

\noindent\textbf{Lemma 2 (Matching Probability).}
Under Assumption~3 and the approximate independence of keypoints,
\begin{equation}
    \mathbb{E}[\eta] 
    \;\le\; \exp(-\alpha \beta_r \Delta),
\end{equation}
where $\eta$ is the inlier ratio.  
Thus, cross-view gradient inconsistency reduces the expected match rate exponentially.

\paragraph{Impact on SfM Pose Estimation.}
\textbf{Theorem 1 (SfM Breakdown Condition).}
Consider a pose graph whose spanning tree contains $m$ critical edges.  
If each critical view pair is poisoned such that
\[
\mathcal{L}_{\mathrm{CVI}} 
\ge \tau_g + \frac{\tau_d}{\beta_r} + \Delta,
\]
and each view contributes at least $N$ keypoints, then the global SfM optimization fails with probability
\begin{equation}
\begin{split}
\Pr[\text{SfM fail}] 
&\ge 1 - m \exp\!\left(
   -\tfrac{1}{2}\epsilon_c^2 N p_{\text{match}}
 \right),\\[2pt]
p_{\text{match}} 
&= \exp(-\alpha \beta_r \Delta),
\end{split}
\end{equation}
where $\epsilon_c$ ensures $(1-\epsilon_c)Np_{\text{match}} < \eta_{\min}$, the minimum inlier threshold for accepting an edge in RANSAC.

\vspace{0.2em}
\noindent\textit{Interpretation.}  
Once each critical view pair exceeds the inconsistency threshold 
\[
L_{\mathrm{th}}=\tau_g+\tau_d/\beta_r,
\]
the matching probability drops sharply, causing critical pose-graph edges to fail and leaving the SfM system under-constrained.  
This provides a stylized yet physically meaningful explanation of why increasing $\texttt{CVIPLoss}$ leads to global reconstruction collapse.
We validate all theoretical thresholds empirically in Appendix~\ref{sup:exp:experimental_validation_of_the_theory}.

\subsection{Proxy-Guided Optimization}

Our method leverages the fact that multi-view consistency naturally emerges from the image formation process. To obtain gradients that are otherwise inaccessible from non-differentiable SfM pipelines such as COLMAP, we train a differentiable 3DGS~\cite{kerbl3Dgaussians} proxy that approximates the victim model's cross-view behavior. This proxy renders multi-view images that reveal how small perturbations influence geometric consistency across views.

Following the theoretical formulation of $\texttt{CVIPLoss}$ in~\Cref{sec:method:theory}, we optimize only the poisoned-view subset $\mathcal{A}$ using projected gradient ascent, while the clean views $\mathcal{R}$ remain unchanged and serve as reference geometry. At each iteration, the proxy renderer $P$ evaluates cross-view inconsistency between $\mathcal{A}$ and $\mathcal{R}$ and produces differentiable gradients for $\texttt{CVIPLoss}$. These gradients are combined with SSIM and TV regularizers to preserve visual fidelity. Each poisoned view is then updated via $\ell_\infty$-bounded PGD:
\begin{equation}
    \tilde I_k \leftarrow 
    \mathrm{Proj}_{\|\tilde I_k - I_k\|_\infty \le \varepsilon}
    \big( \tilde I_k + \alpha \,
    \mathrm{sign}(\nabla_{\tilde I_k} L) \big),
    \quad k \in \mathcal{A}.
\end{equation}

Because $\texttt{CVIPLoss}$ depends on view-dependent visibility, the proxy is periodically refreshed every $K$ iterations to ensure reliable cross-view gradients as the poisoned images evolve. This results in an alternating optimization scheme in which the inner loop updates the poisoned views, while a lightweight outer update keeps the proxy aligned with changing appearance and visibility. Although the theoretical threshold $L_{\mathrm{th}}$ predicts when SfM failure becomes likely, we simply optimize for a fixed number of steps $T$, as guided by the theoretical analysis in~\Cref{app:theory}.

This procedure yields visually imperceptible poisoned views that systematically break the multi-view consistency required for SfM initialization. Consequently, the attack destabilizes the early stages of reconstruction and compromises all downstream 3D pipelines. The complete framework is summarized in Algorithm~\ref{alg:pointview}.

\begin{algorithm}[!t]
\caption{Proxy-guided PGD for \name}
\label{alg:pointview}
\begin{algorithmic}[1]
\STATE {\bf Input:} Clean images $\{I_k\}$, budget $\varepsilon$, steps $T$, step size $\alpha$, proxy $P(\cdot)$
\STATE {\bf Output:} Poisoned images $\{\tilde{I}_k\}$
\STATE Initialize $\tilde{I}_k \leftarrow I_k$ for all $k$
\FOR{$t = 1$ to $T$}
  \STATE Compute proxy loss $L_{\mathrm{CVI}} \leftarrow P(\{\tilde{I}_k\})$
  \STATE $L \leftarrow L_{\mathrm{CVI}} - \lambda_{\mathrm{SSIM}}(1-\mathrm{SSIM}(\tilde{I},I)) - \lambda_{\mathrm{TV}}\mathrm{TV}(\tilde{I})$
  \STATE $g \leftarrow \nabla_{\tilde{I}} L$ \COMMENT{via autodiff through proxy}
  \STATE $\tilde{I} \leftarrow \tilde{I} + \alpha\,\mathrm{sign}(g)$
  \STATE $\tilde{I} \leftarrow \mathrm{clip}(\tilde{I}, I-\varepsilon, I+\varepsilon)$
\ENDFOR
\STATE {\bf Return:} $\{\tilde{I}_k\}$
\end{algorithmic}
\end{algorithm}
\section{Experiments}
\label{sec:exp}


\begin{table*}[t]
\centering
\caption{Clean vs. Poisoned comparison across reconstruction pipelines on \textit{T\&T} with $\rho = {16}/{255}$. The proxy model for optimizing the perturbations is 3DGS.
Clean results in black; poisoned results in \textcolor{red!65!black}{parentheses}.}
\label{tab:avg_results}
\resizebox{0.95\textwidth}{!}{
\begin{tabular}{l|ccc|ccc|ccc}
\toprule
\multirow{2}{*}{\textbf{Scene}} &
\multicolumn{3}{c|}{\textbf{Colmap (based on SfM+MVS)}} &
\multicolumn{3}{c|}{\textbf{Instant NGP (based on NeRF)}} &
\multicolumn{3}{c}{\textbf{Mip-Splatting (based on another 3DGS variant)}} \\ 
\cmidrule(lr){2-10}
 & PSNR $\uparrow$ & SSIM $\uparrow$ & LPIPS $\downarrow$&
   PSNR $\uparrow$ & SSIM $\uparrow$ & LPIPS$\downarrow$ &
   PSNR $\uparrow$ & SSIM $\uparrow$ & LPIPS $\downarrow$\\ 
\midrule
Auditorium & 14.55 \poisoned{11.94} & 0.390 \poisoned{0.319} & 0.637 \poisoned{0.687} &
20.67 \poisoned{15.96} & 0.761 \poisoned{0.659} & 0.429 \poisoned{0.514} &
24.41 \poisoned{18.86} & 0.872 \poisoned{0.698} & 0.196 \poisoned{0.352} \\

Courtroom &
14.48 \poisoned{11.64} & 0.371 \poisoned{0.315} & 0.537 \poisoned{0.600} &
19.44 \poisoned{14.85} & 0.640 \poisoned{0.540} & 0.448 \poisoned{0.563} &
23.00 \poisoned{16.12} & 0.791 \poisoned{0.670} & 0.165 \poisoned{0.358} \\

Museum &
14.16 \poisoned{11.45} & 0.418 \poisoned{0.352} & 0.495 \poisoned{0.593} &
15.19 \poisoned{10.34} & 0.471 \poisoned{0.325} & 0.606 \poisoned{0.726} &
20.88 \poisoned{14.63} & 0.768 \poisoned{0.615} & 0.158 \poisoned{0.349} \\

Palace &
9.08 \poisoned{6.99} & 0.407 \poisoned{0.335} & 0.784 \poisoned{0.840} &
19.09 \poisoned{13.48} & 0.668 \poisoned{0.565} & 0.440 \poisoned{0.522} &
19.63 \poisoned{12.99} & 0.731 \poisoned{0.556} & 0.354 \poisoned{0.473} \\

Temple &
7.32 \poisoned{4.42} & 0.408 \poisoned{0.348} & 0.617 \poisoned{0.687} &
17.84 \poisoned{13.58} & 0.689 \poisoned{0.577} & 0.424 \poisoned{0.525} &
20.55 \poisoned{14.86} & 0.805 \poisoned{0.686} & 0.226 \poisoned{0.385} \\

Family &
10.28 \poisoned{7.99} & 0.440 \poisoned{0.383} & 0.550 \poisoned{0.616} &
22.59 \poisoned{16.86} & 0.761 \poisoned{0.644} & 0.235 \poisoned{0.322} &
24.55 \poisoned{18.47} & 0.872 \poisoned{0.696} & 0.095 \poisoned{0.215} \\

Francis &
8.33 \poisoned{6.06} & 0.302 \poisoned{0.239} & 0.686 \poisoned{0.758} &
24.38 \poisoned{18.61} & 0.824 \poisoned{0.715} & 0.265 \poisoned{0.373} &
27.61 \poisoned{21.33} & 0.899 \poisoned{0.725} & 0.172 \poisoned{0.355} \\

Horse &
6.35 \poisoned{3.72} & 0.404 \poisoned{0.327} & 0.563 \poisoned{0.613} &
21.82 \poisoned{17.31} & 0.784 \poisoned{0.651} & 0.225 \poisoned{0.316} &
23.94 \poisoned{17.70} & 0.879 \poisoned{0.755} & 0.104 \poisoned{0.278} \\

Lighthouse &
10.68 \poisoned{8.45} & 0.514 \poisoned{0.457} & 0.665 \poisoned{0.745} &
21.65 \poisoned{16.06} & 0.765 \poisoned{0.629} & 0.281 \poisoned{0.368} &
22.25 \poisoned{16.45} & 0.844 \poisoned{0.723} & 0.159 \poisoned{0.330} \\

M60 &
14.19 \poisoned{12.18} & 0.547 \poisoned{0.471} & 0.550 \poisoned{0.651} &
25.82 \poisoned{19.82} & 0.832 \poisoned{0.703} & 0.202 \poisoned{0.321} &
27.98 \poisoned{21.31} & 0.904 \poisoned{0.749} & 0.112 \poisoned{0.311} \\

Panther &
14.78 \poisoned{12.74} & 0.594 \poisoned{0.538} & 0.521 \poisoned{0.654} &
28.32 \poisoned{22.38} & 0.908 \poisoned{0.774} & 0.151 \poisoned{0.252} &
28.27 \poisoned{21.57} & 0.908 \poisoned{0.756} & 0.109 \poisoned{0.252} \\

Playground &
13.30 \poisoned{10.97} & 0.455 \poisoned{0.384} & 0.615 \poisoned{0.697} &
23.33 \poisoned{17.49} & 0.696 \poisoned{0.593} & 0.344 \poisoned{0.427} &
25.87 \poisoned{20.37} & 0.861 \poisoned{0.684} & 0.155 \poisoned{0.290} \\

Courthouse &
9.09 \poisoned{6.63} & 0.397 \poisoned{0.323} & 0.737 \poisoned{0.793} &
20.80 \poisoned{15.14} & 0.581 \poisoned{0.555} & 0.414 \poisoned{0.543} &
22.15 \poisoned{16.37} & 0.779 \poisoned{0.616} & 0.265 \poisoned{0.392} \\

Train &
11.83 \poisoned{9.36} & 0.396 \poisoned{0.313} & 0.813 \poisoned{0.894} &
21.00 \poisoned{15.45} & 0.658 \poisoned{0.554} & 0.344 \poisoned{0.546} &
21.82 \poisoned{15.49} & 0.795 \poisoned{0.632} & 0.172 \poisoned{0.336} \\

Truck &
13.35 \poisoned{11.05} & 0.507 \poisoned{0.445} & 0.545 \poisoned{0.646} &
22.85 \poisoned{16.85} & 0.770 \poisoned{0.623} & 0.216 \poisoned{0.443} &
24.36 \poisoned{19.51} & 0.857 \poisoned{0.696} & 0.108 \poisoned{0.236} \\

\midrule
\textbf{Average} &
11.92 \poisoned{8.96} & 0.436 \poisoned{0.372} & 0.606 \poisoned{0.693} &
21.62 \poisoned{16.24} & 0.712 \poisoned{0.605} & 0.340 \poisoned{0.440} &
23.93 \poisoned{17.63} & 0.833 \poisoned{0.684} & 0.166 \poisoned{0.327} \\
\midrule
\textbf{Avg. Drop (\%)} & \textcolor{red!65!black}{-24.8 \%} & \textcolor{red!65!black}{-14.7 \%} & \textcolor{red!65!black}{+14.4 \%} &
\textcolor{red!65!black}{-24.8 \%} & \textcolor{red!65!black}{-15.0 \%} & \textcolor{red!65!black}{+29.4 \%} &
\textcolor{red!65!black}{-26.0
 \%} & \textcolor{red!65!black}{-18.4 \%} & \textcolor{red!65!black}{+92.4 \%} \\
\bottomrule
\end{tabular}}
\end{table*}

\subsection{Experimental Setups}

\paragraph{Datasets.}
We evaluate our method on three standard benchmarks for 3D reconstruction.
(1) \emph{NeRF-Synthetic}~\citep{mildenhall2021nerf}\footnote{\url{https://github.com/bmild/nerf}} provides clean synthetic scenes with ground-truth geometry and camera poses, serving as a controlled environment for evaluating reconstruction quality.
(2) \emph{Tanks and Temples (T\&T)}~\citep{Knapitsch2017}\footnote{\url{https://www.tanksandtemples.org/download/}}
offers high-resolution real-world images with rich textures, occlusions, and realistic geometry, enabling evaluation under practical conditions.
(3) \emph{Mip-NeRF360}~\citep{barron2022mipnerf360}\footnote{\url{https://jonbarron.info/mipnerf360/}} contains 9 outdoor and 4 indoor real-world scenes featuring complex central objects and detailed backgrounds.

\paragraph{Evaluation Metrics.}
We evaluate both the internal SfM statistics and the downstream rendering quality to assess the impact of poisoning. 
For SfM, we report three indicators: 
\emph{Registered Images}, measuring the success rate of camera registration and overall pipeline stability; 
\emph{Triangulated Keypoints}, reflecting the number of successfully matched and triangulated features as a measure of geometric consistency; 
and \emph{Total 3D Points}, representing the density of the reconstructed structure and its completeness. 
For downstream evaluation, following prior work~\citep{barron2022mipnerf360, kerbl3Dgaussians, mildenhall2021nerf}, we use three perceptual metrics: PSNR, SSIM~\citep{hore2010image}, and LPIPS~\citep{zhang2018unreasonable}, to capture pixel accuracy, structural similarity, and perceptual realism of the reconstructed views.

\paragraph{Implementation Details.}
Experiments are conducted on an NVIDIA A100 (40GB). 
We adopt a two-level PGD-based poisoning procedure. 
The perturbation budget is $\rho = 16/255$ under the $\ell_\infty$ constraint. 
Each adversarial view is optimized for $15$ PGD steps with step size $\alpha = 2/255$, using random initialization within the $\ell_\infty$ ball. 
The outer optimization runs for $1000$ iterations, during which adversarial views are injected with a poisoning ratio of $r=0.6$. 
The \texttt{CVIPLoss} combines gradient inconsistency ($w_{\text{grad}}=1.0$), total variation ($w_{\text{tv}}=0.1$), and SSIM regularization ($w_{\text{ssim}}=0.5$). 
The proxy is refreshed every K=10 iterations in all experiments unless otherwise specified.
Other reconstruction settings follow the defaults of COLMAP~\cite{schoenberger2016sfm, schonberger2016pixelwise}, Instant NGP~\cite{mueller2022instant}, 3D Gaussian Splatting~\cite{kerbl3Dgaussians}, and Mip-Splatting~\cite{Yu2023MipSplatting}.
We select 3DGS as the proxy in all experiments, as its high efficiency enables fast optimization and large-scale evaluation.

\paragraph{Baselines.}
To investigate the effectiveness of our method.
We experiment with three types of strategy:
(1) \textbf{Gauss. noise}, i.e., directly add Gaussian noise to input views, (2) \textbf{Single-view methods}, i.e., directly push the rendered views away from the ground truth views (without cross-view considerations)~\cite{lin2023nerfool,meng2025il2nerf}, and (3) \textbf{Structural Upper Bound}: To estimate the upper bound of our method on SfM statistics, we manually mask keypoints and remove all pixel-level structural features. Details are provided in Appendix~\ref{sup:exp}.

\begin{table}[t]
\centering
\small  
\setlength{\tabcolsep}{6pt} 
\renewcommand{\arraystretch}{0.5}
\caption{
Our \name~compared to other baselines. Metrics are averaged over all scenes on \emph{T\&T} dataset with the Mip-Splatting~\cite{Yu2023MipSplatting}. }
\label{tab:baselines}
\begin{tabular}{lcc}
\toprule
{Baselines}
& PSNR & SSIM \\
\midrule
Clean &23.93& 0.833 \\
\midrule
Gauss. noise &23.43 & 0.821\\
Single-view~\cite{lin2023nerfool,meng2025il2nerf} & 23.53 & 0.819 \\
Ours (Cross-view)           & \textbf{17.63} & \textbf{0.684}  \\
\bottomrule
\vspace{-3mm}
\end{tabular}
\end{table}

\subsection{Quantitative Results}
\paragraph{Downstream Reconstruction Quality.}
As shown in ~\Cref{tab:avg_results}, our attack consistently degrades all three 3D reconstruction pipelines under the same perturbation budget ($\rho=16/255$). 
On \emph{T\&T} dataset, PSNR decreases by about 25\%, SSIM by 15–18\%, and LPIPS nearly doubles on average. 
The uniform degradation across COLMAP, Instant NGP, and 3D Gaussian Splatting demonstrates the downstream-agnostic nature of the attack. 
Among them, 3D Gaussian Splatting shows the largest LPIPS increase, suggesting that perturbations injected before SfM are amplified through its geometry-dependent rendering. 
These results confirm that corrupting SfM initialization fundamentally destabilizes all subsequent reconstruction stages, regardless of the underlying 3D representation.
All clean baselines for MVS, NeRF, and 3DGS are obtained from the NeRF Baselines repository~\cite{kulhanek2025nerfbaselines}\footnote{\url{https://nerfbaselines.github.io}}.
As shown in \Cref{tab:baselines}, our \name~achieves substantially better results than the noise-based and single-view-based attacks.
More quantitative results related to other datasets, see Appendix~\ref{sup:exp:quantitative_results}.

\paragraph{SfM Stability and Geometric Collapse.}
Table~\ref{tab:sfm_stability} summarizes SfM stability across three datasets.
Compared with random or structural noise, our poisoning attack causes a sharp decline in registered views, triangulated keypoints, and sparse 3D points, yielding an average collapse ratio above 70\%.

\begin{table}[t]
\centering
\renewcommand{\arraystretch}{0.7}
\caption{\textbf{Average performance across three datasets.}
Values are averaged over all scenes with Mip-Splatting~\cite{Yu2023MipSplatting}.
Reg.(\%) denotes registered view ratio, Triang.(k) denotes triangulated keypoints ($\times 10^{3}$), and 3D Pts(M) denotes sparse 3D points ($\times 10^{6}$).
}
\label{tab:sfm_stability}
\resizebox{0.95\linewidth}{!}{
\begin{tabular}{lccccc}
\toprule
\multirow{2}{*}{\textbf{Dataset}} &
\multirow{2}{*}{\textbf{Attack}} &
\multicolumn{3}{c}{\textbf{SfM Stability}} &
\multirow{2}{*}{\shortstack{\textbf{Collapse}\\\textbf{Ratio}}} \\
\cmidrule(lr){3-5}
& & Reg.(\%) & Triang.(k) & 3D Pts(M) & \\
\midrule
\multirow{4}{*}{NeRF-Synthetic}
 & Clean & 98.7 & 52.3 & 2.11 & 0.00 \\
 & Gauss. noise & 97.5 & 50.9 & 2.03 & 0.02 \\
 & Structural UB  & 25.7 & 11.2 & 0.28 & 0.86 \\
 & \textbf{Ours} & \textbf{28.5} & \textbf{12.4} & \textbf{0.32} & \textbf{0.83} \\
\midrule
\multirow{4}{*}{Mip-NeRF360}
 & Clean & 96.2 & 61.7 & 2.64 & 0.00 \\
 & Gauss. noise & 94.8 & 58.5 & 2.49 & 0.04 \\
 & Structural UB  & 24.5 & 10.9 & 0.30 & 0.87 \\
 & \textbf{Ours} & \textbf{26.9} & \textbf{11.8} & \textbf{0.34} & \textbf{0.85} \\
\midrule
\multirow{4}{*}{Tanks \& Temples}
 & Clean & 93.5 & 73.6 & 3.07 & 0.00 \\
 & Gauss. noise & 91.2 & 70.8 & 2.95 & 0.05 \\
 & Structural UB & 22.8 & 12.1 & 0.36 & 0.90 \\
 & \textbf{Ours} & \textbf{24.3} & \textbf{13.7} & \textbf{0.41} & \textbf{0.88} \\
\bottomrule
\end{tabular}}
\end{table}

\begin{figure*}[htbp]
    \centering
    \includegraphics[width=0.85\linewidth]{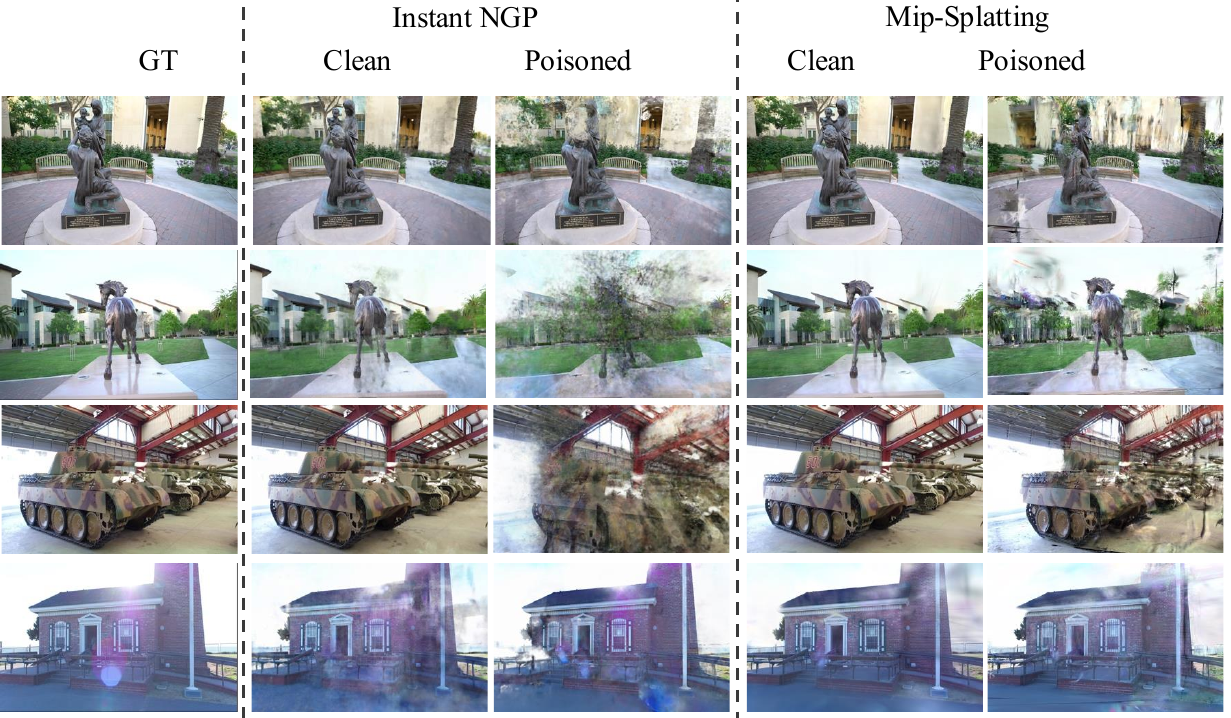}
    \caption{Qualitative comparison of clean and poisoned reconstructions across multiple scenes on \emph{T\&T}. From left to right: ground-truth image, clean reconstruction, and poisoned reconstruction for Instant NGP~\cite{mueller2022instant} and Mip-Splatting~\cite{Yu2023MipSplatting}.
    While clean inputs reproduce accurate geometry and appearance, poisoned inputs lead to severe structural collapse, view-inconsistent artifacts, and distorted scene geometry. The consistent failure across both pipelines demonstrates that our cross-view poisoning destabilizes the underlying 3D structure required for reliable reconstruction.
    }
    \label{fig:visualization_for_poisoned}
\end{figure*}



\subsection{Qualitative Results}
\begin{figure}[!t]
    \centering
    \includegraphics[width=\linewidth]{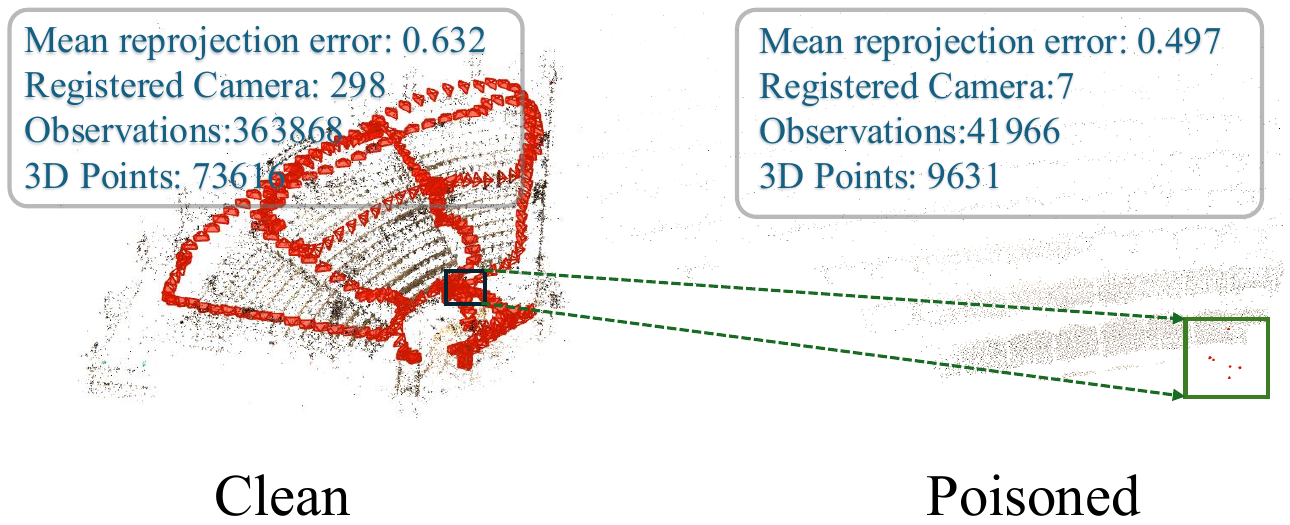}
    \caption{
    Clean vs. poisoned COLMAP reconstructions on the ``Auditorium'' scene of \emph{T\&T}. 
    The attack collapses the geometry (298→7 cameras, 73k→9k points), even though the reprojection error appears lower. 
    Green box: remaining registered cameras.
}
    \label{fig:Clean_and_Poisoned_comparison}
\end{figure}

\Cref{fig:visualization_for_poisoned} shows that clean inputs yield reconstructions close to the ground truth for both Instant NGP and Gaussian Splatting.
Poisoned inputs, however, cause drastic geometric collapse and view-inconsistent artifacts across all scenes.
The consistent failure across two fundamentally different pipelines demonstrates the robustness and generality of our attack.
~\Cref{fig:Clean_and_Poisoned_comparison} illustrates this effect on the \emph{Auditorium} scene: 
The poisoned reconstruction collapses almost entirely, with only a handful of views surviving.
Despite appearing to have a lower reprojection error, the structure is largely unrecoverable,
showing that cross-view inconsistencies fundamentally disrupt SfM initialization.

\subsection{Stealthiness of \name}
\begin{figure}[t]
    \centering
    \includegraphics[width=\linewidth]{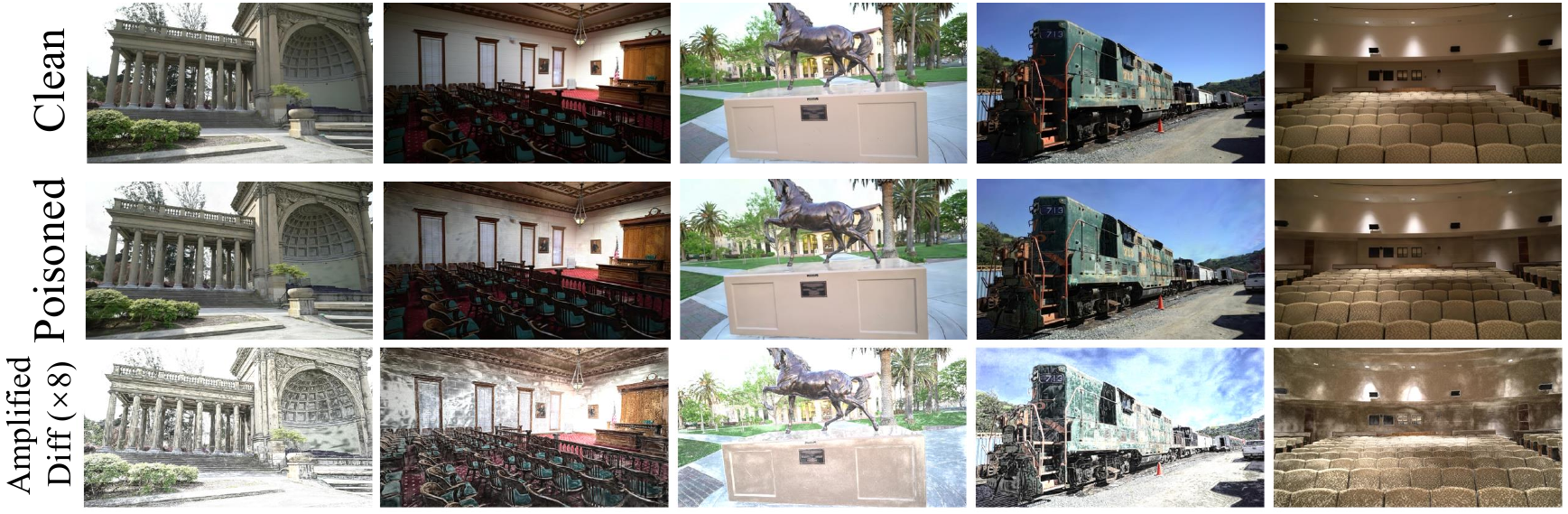}
    \caption{
    Stealthiness of PoInit-of-View on \emph{T\&T}. Clean and poisoned images look similar, while amplified difference maps ($\times 8$) reveal subtle perturbations.
    }
    \label{fig:vis:perturbation}
\end{figure}

\begin{table}[t]
\centering
\caption{Perceptual similarity between clean and poisoned images (mean ± std over 5 scenes).}
\scriptsize
\renewcommand{\arraystretch}{0.8}  
\begin{tabular}{ccc}
\toprule
PSNR (dB) $\uparrow$ & SSIM $\uparrow$ & LPIPS $\downarrow$ \\
\midrule
$27.8 \pm 2.6$ & $0.86 \pm 0.06$ & $0.21 \pm 0.04$ \\
\bottomrule
\end{tabular}
\renewcommand{\arraystretch}{1.0} 
\label{tab:stealth}
\end{table}

~\Cref{fig:vis:perturbation} visualizes the clean, poisoned, and amplified-difference views to qualitatively assess the subtlety of the perturbations. 
To assess perceptual imperceptibility more quantitatively, we measure PSNR, SSIM~\cite{hore2010image}, and LPIPS~\cite{zhang2018unreasonable} 
between the clean and poisoned views (\Cref{tab:stealth}). 
The results indicate that the perturbations are visually imperceptible. 
Despite such subtle pixel-level changes, SfM registration collapses by over 80\%, 
demonstrating that even minor cross-view inconsistencies are sufficient to disrupt geometric initialization.

\subsection{Ablations}

\begin{table}[t]
\centering
\renewcommand{\arraystretch}{0.8}
\caption{
Metrics are averaged over all scenes on \emph{T\&T} dataset with the Mip-Splatting~\cite{Yu2023MipSplatting}. }
\label{tab:proxy_ablation}
\resizebox{\linewidth}{!}{
\begin{tabular}{l|ccc|cc|cc}
\toprule
\multirow{2}{*}{Proxy Design} 
& \multicolumn{3}{c|}{\textbf{Perceptual Similarity}} 
& \multicolumn{2}{c|}{\textbf{SfM Stability}} 
& \multicolumn{2}{c}{\textbf{Downstream}} \\
& PSNR & SSIM & LPIPS 
& Reg.(\%) & 3D pts(K) 
& PSNR & SSIM \\
\midrule
\textbf{Ours (LCVI + SSIM + TV)} 
& 22.3& 0.847 & 0.174
& 24.3 & 13.7  
& 17.63 & 0.684 \\
w/o SSIM (LCVI + TV) 
& 24.8 & 0.865 & 0.192
& 27.1 & 15.8 
& 19.4 & 0.721 \\
w/o TV (LCVI + SSIM) 
& 25.3 & 0.872 & 0.201
& 28.0 & 16.4 
& 19.7 & 0.735 \\
Photometric-only 
& 28.7 & 0.911 & 0.264
& 91.2 & 71.5
& 22.8 & 0.811 \\
\bottomrule
\end{tabular}
}
\end{table}

\paragraph{Ablation on Proxy Objective.}
Table~\ref{tab:proxy_ablation} shows that removing SSIM or TV leads to a mixed change in perturbation quality: although PSNR and SSIM with respect to the clean images become slightly higher, LPIPS becomes worse, indicating reduced perceptual similarity. Meanwhile, the poisoning effect is slightly weakened. The photometric-only variant retains only a per-pixel color loss while removing both SSIM and TV. However, it still fails to collapse SfM, suggesting that minimizing pixel-wise differences alone is insufficient to induce the cross-view inconsistencies required to disrupt multi-view geometric reconstruction.

\begin{figure}[htbp]
    \centering
    \includegraphics[width=\linewidth]{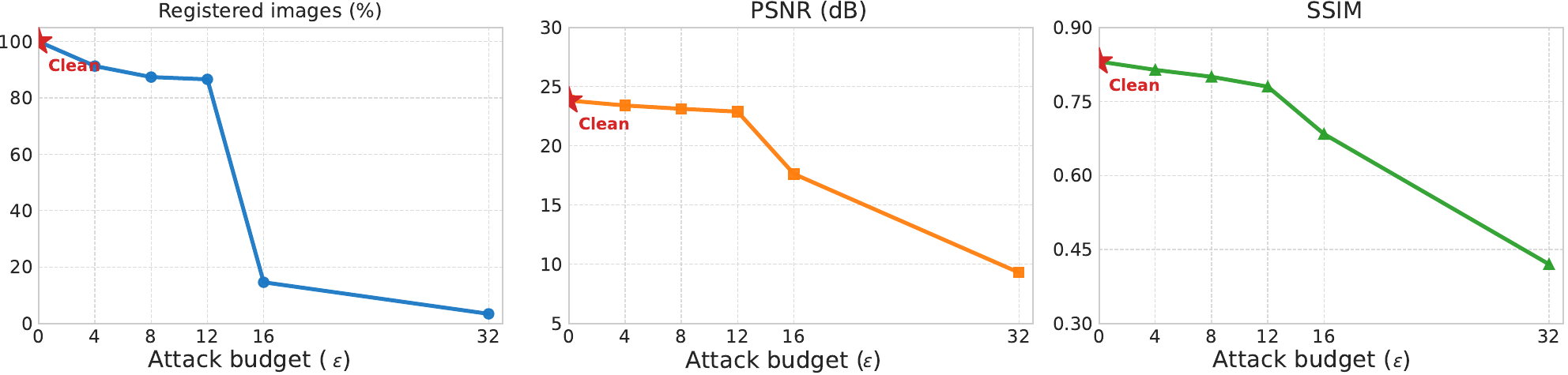}
    \caption{
    Ablation on perturbation budget $\varepsilon$, showing that larger perturbations cause greater degradation in registration and reconstruction.
    }
    \label{fig:epsilon_ablation_metrics}
    \vspace{-3mm}
\end{figure}

\paragraph{Ablation on Perturbation Budget $\rho$.}
As shown in~\Cref{fig:epsilon_ablation_metrics}, increasing the perturbation budget $\rho$ steadily strengthens the attack. Small perturbations have a limited effect, but performance drops sharply once $\rho$ exceeds $12/255$: the registration rate collapses, and both PSNR and SSIM fall rapidly. At $\rho = 32/255$, the pipeline almost completely fails, indicating that the reconstruction becomes highly fragile under moderate-to-large perturbations.

\begin{figure}[htbp]
    \centering
    \includegraphics[width=0.9\linewidth]{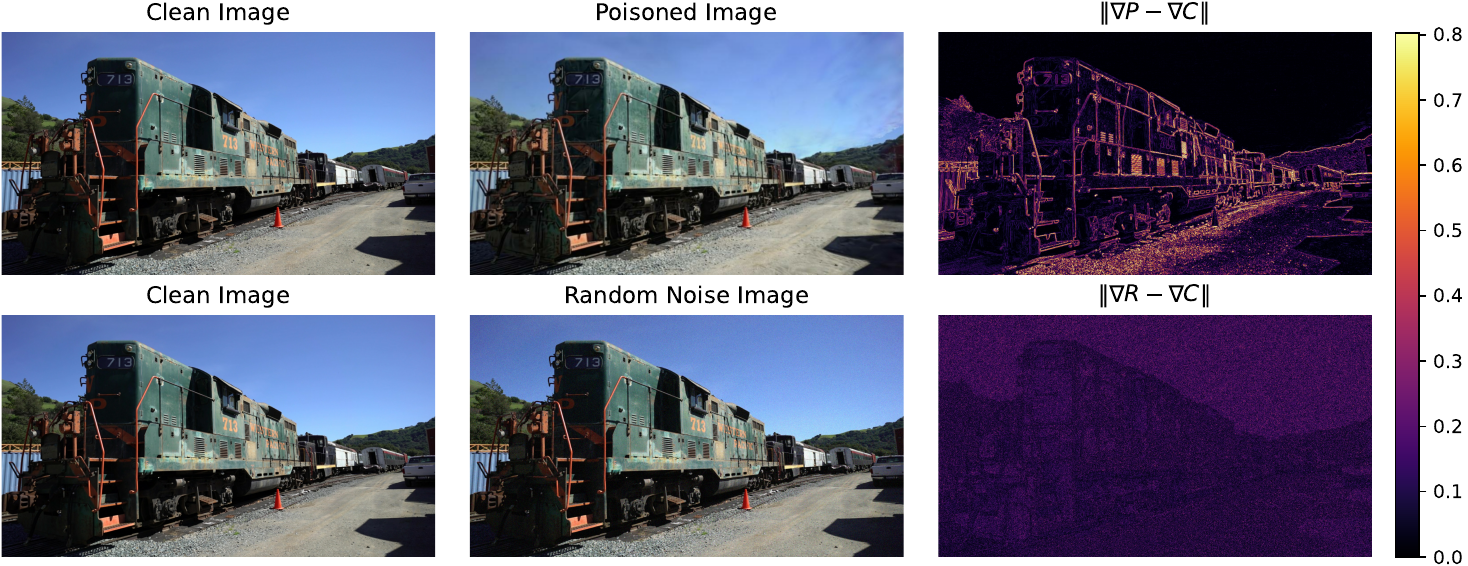}
    \caption{
    Gradient-difference comparison on \emph{T\&T}. 
    Our perturbation remains imperceptible while inducing clear edge-aligned discrepancies, unlike random noise, indicating targeted disruption of SfM-relevant structures.
    }
    \label{fig:structred_perturbations}
\end{figure}

\paragraph{Effect of Structured Perturbations \textit{vs.} Random Noise.}
We compute gradient-difference maps using the Sobel~\cite{kittler1983accuracy} operator between the clean and poisoned images for visualization, as shown in ~\Cref{fig:structred_perturbations}. 
It reveals that our poisoning introduces structured, geometry-aligned perturbations rather than random fluctuations. 
These structured perturbations vary across views and consequently induce cross-view inconsistencies that disrupt the feature correspondence on which SfM relies. 
These effects weaken both geometric cues and multi-view consistency, causing SfM initialization to fail and triggering downstream reconstruction collapse.

\paragraph{Verifying the Theoretical Prediction.}
\begin{figure}[htbp]
    \centering
    \includegraphics[width=\linewidth]{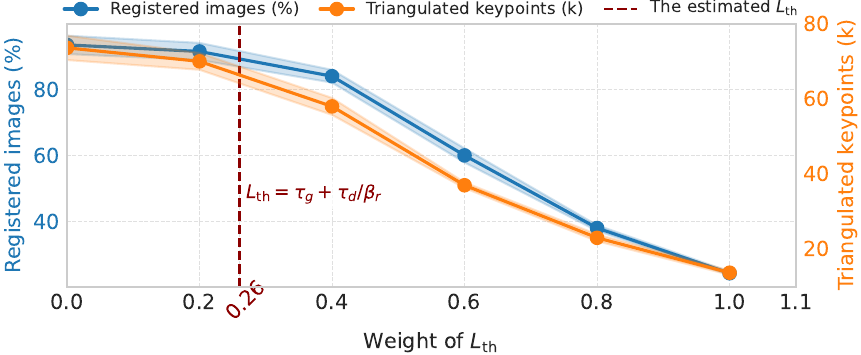}
    \caption{Impact of cross-view inconsistency on SfM. 
    Increasing $\mathcal{L}_{\text{CVI}}$ reduces registered images and triangulated keypoints, with both collapsing at large inconsistency. }
    \label{fig:lcvi_vs_sfm}
    \vspace{-5mm}
\end{figure}

~\Cref{fig:lcvi_vs_sfm} shows that increasing $\mathcal{L}_{\text{CVI}}$ progressively degrades SfM and eventually causes collapse, consistent with our theoretical analysis in ~\Cref{sec:method:theory} on \emph{T\&T}. 
To validate the predicted instability threshold, we empirically estimate $(\tau_g, \tau_d, \beta_r)$ on clean SfM reconstructions and compute the corresponding $L_{\text{th}}$, which is plotted as the vertical dashed line in~\Cref{fig:lcvi_vs_sfm}.

The full estimation procedure and the numerical value of $L_{\text{th}}$ are provided in Appendix~\Cref{sup:exp:experimental_validation_of_the_theory}, which can empirically validate the theoretical breakdown condition.
Further ablations of other hyperparameters, along with related discussions of defenses, collapse ratio, and cross-model generalizability, are provided in Appendix~\Cref{sup:exp}.

\section{Conclusion and Outlook}
We presented \name, a poisoning attack that exposes a fundamental vulnerability of the structure-from-motion (SfM) initialization model in modern 3D reconstruction pipelines. 
We also provided a theoretical analysis linking cross-view inconsistency to correspondence collapse in SfM. 
Extensive experiments on multiple benchmarks verified the effectiveness of our \name~and its superiority over existing single-view-based attacks.
Future work may investigate attack transferability across heterogeneous SfM implementations. 
Another promising direction lies in developing defenses for multi-view geometry, such as robust feature matching and pose optimization with adversarial regularization.


\section{Acknowledgement}
This work was supported by the New Generation Artificial Intelligence-National Science and Technology Major Project (2025ZD0123305), the National Natural Science Foundation of China (U244120060, 62406240), EU Horizon projects TANGO (No. 101120763) and ELIAS (No. 101120237), and the FIS project GUIDANCE (No. FIS2023-03251).

{
    \small
    \bibliographystyle{ieeenat_fullname}
    \bibliography{main}
}


\end{document}